\documentclass{article}
\usepackage{spconf,amsmath,graphicx}
\usepackage{booktabs}
\usepackage{kbordermatrix}
\usepackage{amsfonts}
\usepackage{float}

\DeclareMathOperator*{\argmax}{arg\,max}

\title{Multi-view Dimensionality Reduction for Dialect Identification of Arabic Broadcast Speech}
\twoauthors
  {Sameer Khurana, Ahmed Ali}
	{Qatar Computing Research Institute\\ HBKU, Doha, Qatar\\}
  {Steve Renals}
	{Centre for Speech Technology Research\\ University of Edinburgh, UK\\}
\begin{document}
%
\maketitle
\begin{abstract}
In this work, we present a new Vector Space Model (VSM) of speech utterances for the task of spoken dialect identification. Generally, DID systems are built using two sets of features that are extracted from speech utterances; acoustic and phonetic. The acoustic and phonetic features are used to form vector representations of speech utterances in an attempt to encode information about the spoken dialects. The Phonotactic and Acoustic VSMs, thus formed, are used for the task of DID. The aim of this paper is to construct a single VSM that encodes information about spoken dialects from both the Phonotactic and Acoustic VSMs. Given the two views of the data, we make use of a well known multi-view dimensionality reduction technique known as Canonical Correlation Analysis (CCA), to form a single vector representation for each speech utterance that encodes dialect specific discriminative information from both the phonetic and acoustic representations. We refer to this approach as feature space combination approach and show that our CCA based feature vector representation performs better on the Arabic DID task than the phonetic and acoustic feature representations used alone. We also present the feature space combination approach as a viable alternative to the model based combination approach, where two DID systems are built using the two VSMs (Phonotactic and Acoustic) and the final prediction score is the output score combination from the two systems.

\end{abstract}
\begin{keywords}
Canonical Correlation Analysis (CCA), Multi-view Dimensionality Reduction, Vector Space Model (VSM), Arabic Dialect Identification (DID)
\end{keywords}
\section{Introduction}
\label{sec:intro}
Dialect Identification (DID) problem is a special case of the more general problem of Language Identification (LID). LID refers to the process of automatically identifying the language class for given speech segment or text document, while DID classifies between dialects within the same language class, making it a more challenging task than LID. A good DID system used as a front-end to an automatic speech recognition system, can help improve the recognition performance by providing dialectal data for acoustic and language model adaptation  to the specific dialect being spoken \cite{biadsy2011automatic}.

In this work, we focus on Arabic DID which can can be posed as a five class classification problem, given that the Arabic language can be divided into five major dialects; Egyptian (EGY), Gulf (GLF), Lavantine (LAV), Modern Standard Arabic (MSA) and North African (NOR) \cite{ali2016did}.

Over the past decade, great advances have been made in the field of automatic language identification (LID). Research effort has focused on coming up with mathematical representations of speech utterances, that encodes the information about the language being spoken. These approaches are also known as Vector Space Modeling approaches \cite{li2007vector}, where speech utterances are represented by a continuous vector of high dimensions. Two predominant Vector Space Modeling approaches are Phonotactic and Acoustic. \textbf{Phonotactic approaches} attempt to model the n-gram phone statistics of speech. Phone sequences for each utterance are extracted using one or multiple phone recognisers. A Vector Space Model (VSM) is then constructed using a \textit{term-document matrix} \cite{deerwester1990indexing}, followed by an unsupervised dimensionality reduction technique, such as Principal Component Analysis (PCA) \cite{moore1981principal} to map the high dimensional feature space to a low dimensional Vector Subspace (Section~\ref{sec:pvsm}), giving a Phonotactic VSM. In other cases, a phone n-gram language model is used to model the phone statistics instead of a VSM \cite{zissman2001automatic,biadsy2009spoken,biadsy2011dialect}. On the other hand, \textbf{Acoustic approaches} attempt to extract dialect discriminative information from speech using low level acoustic features, such as pitch, prosody, shifted delta ceptral coefficients, bottleneck features \cite{torresapproaches,martinez2012ivector}. One of the most successful acoustic approaches is, the use of i-Vector framework for LID, where i-Vectors are extracted for each speech utterance, using an i-Vector extractor that consists of a GMM-UBM trained on top of BNF, followed by a \textit{Total Variability Subspace Model} \cite{ali2016did,dehak2011language}. The extracted i-Vectors give an Acoustic VSM (Section~\ref{sec:avsm}). These methods are also used for DID.

Each of the two VSMs is used as an input to a back-end discriminative classifier, which is trained to find a suitable decision boundary in these vector spaces. This gives us two DID systems built using the Acoustic and Phonotactic VSMs. At prediction time, output scores from the two DID systems are combined to give a final score, on the basis of which classification decision is made. This model combination approach has been shown to give performace improvements on the DID task \cite{ali2016did}. This also shows that the two systems are complementary to each other, which leads us to investigate a feature space combination approach i.e. to construct a single VSM by combining Phonotactic and Acoustic VSMs, in an attempt to encode useful discriminative information in that single VSM.

In this work, we present a \textbf{feature space combination} approach. We form a combined VSM that incorporates useful information, necessary for DID, from both the Phonotactic and Acoustic VSMs. To achieve this goal, we make use of the well known multi-view dimensionality reduction technique known as Canonical Correlation Analysis (CCA), devloped by H. Hotelling \cite{ccahoteling} (Section~\ref{sec:ccavsm}). We show the performance of the combined VSM on Arabic DID task and compare it against the performance of Phonotactic and Acoustic VSMs used alone (Section~\ref{sec:exp}). CCA VSM shows superior performance. The advantages of our feature space combination approach over model combination are two fold: Only one back-end classifier needs to be trained and; Unlabeled data from other domain can easily be used in CCA framework to construct the single VSM. In this work, we do not experiment with unlabeled data and leave it as an extension to our current work.

\section{Vector Space Models}
\label{sec:vsm}
This section gives details about the construction of combined VSM, also referred to as CCA VSM, $\mathbf{Z_{C}}$. We start by presenting the Phonotactic VSM, $X_{P}$ and Acoustic VSM, $X_{A}$, used in this work, followed by the section on CCA VSM, $Z_{C}$. 

\subsection{Phonotactic VSM; $\mathbf{X_P}$}
\label{sec:pvsm}
Phonotactic VSM is constructed by modeling the n-gram phone statistics of the phone sequences that are extracted using an Arabic phone recognizer. Details about the phone recognizer can be found in \cite{ali2016did}. VSM is constructed in two steps; 1) Construct a \textit{term-document matrix}, $\mathbf{X} \in \mathbb{R}^{N \times d}$ (See Fig~\ref{pfs}), where each speech utterance in represented by a Phonotactic feature vector, $\mathbf{p} = \left(f(p,s_1), f(p,s_2), \ldots, f(p, s_d)\right) \in \mathbb{R}^{d\times 1}$, where $N$ is the number of speech utterances and $f(p,s)$ is the number of times a phone n-gram (term) $s$ appears in the utterance (document) $p$ and 2) Perform \textit{Truncated Singular Value Decomposition (SVD)} (Equation~\ref{eq:1}) on $\mathbf{X}$ to learn a lower dimensional linear manifold, $\mathbf{\Pi} \in \mathbb{R}^{d \times k}$, where $k << d$. SVD attempts to discover the latent structure in the high dimensional feature space. Note that, $k$ is the number of largest singular values. $X$ is projected down to $\Pi$ to get the Phonotactic VSM, $\mathbf{X_{P}}$ (Equation~\ref{eq:2}).

\begin{figure}[htb]
\centering
\small
\[
  \text{$\mathbf{X}$} = \kbordermatrix{
    & s_1 & s_2 & \dots & s_d \\
    p_1 & f(p_1,s_1) & f(p_1,s_2) & \ldots & f(p_1,s_d) \\
    p_2 & f(p_2,s_1) & f(p_2,s_2) & \dots & f(p_2,s_d) \\
    \vdots & \vdots  & \vdots &\ddots & \vdots \\
    p_N & f(p_N,s_1) & f(p_N,s_2) & \dots  & f(p_N,s_d)
  }
\]
\normalsize
\caption{\textit{n-gram phonetic feature space, $\mathbf{X}$}}
\label{pfs}
\end{figure}

\begin{eqnarray}
	\mathbf{\underbrace{\mathbf{X}}_{N\times d}} &=& \mathbf{\underbrace{\mathbf{U}}_{N\times k}\bullet \underbrace{\mathbf{S}}_{k\times k} \bullet \underbrace{\mathbf{\Pi}^{T}}_{k\times d}}\\ \label{eq:1}
	\mathbf{\underbrace{X_{P}}_{N\times k}} &=& \mathbf{\underbrace{X}_{N\times d}} \bullet \mathbf{\underbrace{\Pi}_{d \times k}} \label{eq:2}
\end{eqnarray}

In our case, the n-gram dictionary consisted of phone 2-grams and 3-grams with a total of 8K features i.e. $d=8K$. The dimensionality of the VSM, $X_{P}$, was chosen to be 1200, i.e. $k=1200$. 1200 was the optimal value chosen experimentally.

\subsection{Acoustic VSM; $\mathbf{X_{A}}$}
\label{sec:avsm}

Acoustic VSM is constructed in two steps; 1) Extracting the bottleneck features (BNF) from speech and 2) Modeling BNF using the i-Vector extraction framework. 

We use the same Deep Neural Network (DNN) based ASR system to extract the BNF as in our previous works \cite{ali2016did,cardinal2015speaker}. Two DNNs are used with 5 hidden layers and 1 \textit{Bottleneck Layer}, all having sigmoidal neurons. Tied-phone states are used as the target to the DNNs. The target labels of dimension 3040 are provided by a GMM-HMM baseline system trained on 60 hours of Arabic Broadcast speech \cite{ali2014complete}. Input to the DNN consists of 11 consecutive frames stacked together, where for each frame 23 fbank features along with pitch and voicing probability are extracted. The output of the BN layer from the first DNN are fed as inputs to the second DNN, which acts as a correction DNN for the first model. Time offsets at the input layer of the second DNN are -10, -5, 0, 5 and 10, giving an overall context of 31 frames at the input of the second DNN. The BNF from the first DNN are used in the i-Vector modeling framework.

i-Vector modeling framework consists of building a GMM-UBM on a large amount of data using acoustic features (BNF), to model the dialectal feature space. The sufficient statistics of the GMM-UBM give a general idea of the data spread in the high dimensional Vector Space. GMM-UBM mean supervector is updated while adapting it to each utterance. This update information is encoded in a low dimensional latent vector known as an i-Vector. The latent variable model used to extract i-Vector is called \textit{Total Variability Subspace Model} and is given by the equation:
\begin{equation*}
 M = u + Tv
\end{equation*}
where $u$ is GMM-UBM mean supervector. $v$ is the latent vector, known as the $i-Vector$ and $T$ is the lower dimensional Vector Subspace. The parameters of the model are estimated using \textit{Maximum Likelihood} training criterion. For a detailed explanation of i-Vector modeling framework, reader is directed to excellent work in \cite{kenny2008study,dehak2011language}.

In this work, GMM-UBM model has 2048 gaussian components, MFCC features are extracted using a 25 ms window and the i-Vectors are 400 dimensional \cite{ali2016did}.

Finally, we construct the acoustic VSM, $\mathbf{X_A} \in \mathbb{R}^{N\times 400}$, where the $i^{th}$ row is the 400 dimensional i-Vector representation corresponding to the speech utterance, $\mathbf{a}_{i}$. We also perform Linear Discriminant Analysis (LDA) and Within Class Co-variance Normalization (WCCN) on the Acoustic Vector Space, to increase the discriminative strength of the VSM. This method has been shown to improve DID (LID) performance \cite{ali2016did,dehak2011language}.

\subsection{CCA VSM, $\mathbf{Z_{C}}$}
\label{sec:ccavsm}
\subsubsection{Brief Overview; CCA}
\label{subsub:cca}
\begin{figure}[htb]
  \centering
  \includegraphics[width=50mm]{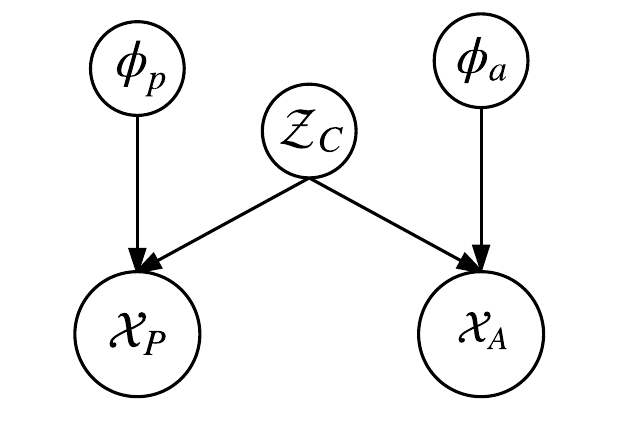}
  \caption{\textit{Graphical Model representation of CCA. Nodes of the graph represent \textit{Random Variables}}}
  \label{ccavsm}
\end{figure}
Here, we give a brief overview of the mathematical foundations of the CCA. Fig~\ref{ccavsm} gives a probabilistic graphical model of CCA. Nodes of the graph represent \textit{Random Variables (RVs)} and the structure encodes conditional independence assumptions. $\mathcal{X_{P}}$ and $\mathcal{X_{A}}$ are the \textit{Random Variables} corresponding to the Phonotactic and Acoustic views of the data. Each data view is associated with two latent variables; 1) $\mathcal{Z_{C}}$, which is shared, and is the variable of interest that will form the final combined VSM, $\mathbf{Z_{C}}$ and 2) $\mathcal{\phi}_{p}$ and $\mathcal{\phi}_{a}$, which are the subspaces associated with the Phontactic and Acoustic views, respectively. CCA attempts to estimate $\phi_{p}$ and $\phi_{a}$ such that the correlation between the projections of the phonotactic feature vectors, $\mathbf{p}$, on $\phi_{p}$ and acoustic feature vectors, $\mathbf{a}$, on $\phi_{a}$ are maximized. Hence, CCA can be posed as the following optimization problem \cite{dhillon2015eigenwords}.
\begin{equation*}
	[\phi_{p},\phi_{a}] = \argmax_{\phi_{p},\phi_{a}} \frac{ \phi^{T}_{p}C_{\mathcal{X}_{P}\mathcal{X}_{A}} \phi_{a}}{\sqrt{\phi^{T}_{p} C_{\mathcal{X}_{P}\mathcal{X}_{P}} \phi_{a}} \sqrt{\phi^{T}_{a} C_{\mathcal{X}_{A}\mathcal{X}_{A}} \phi_{a}}}
\end{equation*}
The above optimization formualtion can be massaged into the following eigenvalue problem. For details see \cite{hardoon2003canonical}.
\begin{eqnarray*}
C^{-1}_{\mathcal{X_{P}}\mathcal{X}_{P}} C_{\mathcal{X}_{P}\mathcal{X}_{A}} C^{-1}_{\mathcal{X}_{A}\mathcal{X}_{A}} C_{\mathcal{X}_{A}\mathbf{X}_{P}} \phi_{p}&=&\lambda \phi_{p}\\
C^{-1}_{\mathcal{X_{A}}\mathcal{X}_{A}} C_{\mathcal{X}_{A}\mathcal{X}_{P}} C^{-1}_{\mathcal{X}_{P}\mathcal{X}_{P}} C_{\mathcal{X}_{P}\mathbf{X}_{A}} \phi_{a}&=&\lambda \phi_{a}
\end{eqnarray*}
An equivalent SVD formulation of the above eigenvalue problem is given below, which allows us to find $\phi_{a}$ and $\phi_{p}$ by performing SVD of $C^{-1/2}_{\mathcal{X}_{P}\mathcal{X}_{P}} C_{\mathcal{X}_{P}\mathcal{X}_{A}} C^{-1/2}_{\mathcal{X}_{A}\mathcal{X}_{A}}$.  For the proof of this formulation, reader is referred to \cite{dhillon2015eigenwords}.
\begin{equation} \label{eq:4}
C^{-1/2}_{\mathcal{X}_{P}\mathcal{X}_{P}} C_{\mathcal{X}_{P}\mathcal{X}_{A}} C^{-1/2}_{\mathcal{X}_{A}\mathcal{X}_{A}} = \phi_{p} \Lambda \phi^{T}_{a}
\end{equation}
We use the above formulation in this paper to learn the latent Vector Subspaces, $\phi_{p}$ and $\phi_{a}$.

\subsubsection{Modeling}
Given the two views, $\mathbf{X_P} \in \mathbb{R}^{N\times 1200}$ and $\mathbf{X_A} \in \mathbb{R}^{N\times 400}$, for our speech data, we form a shared VSM, $\mathbf{Z_C} \in \mathbb{R}^{N\times 2c}$ using CCA formulation discussed in Section~\ref{subsub:cca}. Note that, $\mathbf{X}_{P}$, $\mathbf{X}_{A}$ and $\mathbf{Z}_{C}$ are instantiations of the \textit{RVs} $\mathcal{X}_{P}$, $\mathcal{X}_{A}$ and $\mathcal{Z}_{C}$ respectively as given in Fig~\ref{ccavsm}. $\mathbf{Z}_{C}$'s construction can be concisely given by the following two equations.
\begin{eqnarray}
\left[\mathbf{\underbrace{\phi_p}_{1200\times c}},\ \mathbf{\underbrace{\phi_{a}}_{c\times 400}}\right]&=&\mathbf{CCA}\left[\underbrace{\mathbf{X_{P}}}_{N\times 1200},\ \underbrace{\mathbf{X_{A}}}_{N\times 400}\right]\\
\underbrace{\mathbf{Z_{C}}}_{N\times 2c} &=& \underbrace{\mathbf{X_P \bullet \phi_{p}}}_{N\times c}\Vert \underbrace{\mathbf{X_{A} \bullet \phi^{T}_{a}}}_{N\times c}
\end{eqnarray}
where, $\mathbf{CCA}$ is performed by using the SVD formulation of equation~\ref{eq:4}.

In our case, the shared VSM's dimensionality is 600, i.e. $c=300$. This value is the optimal VSM dimensionality that is experimentally decided. We also perform LDA and WCCN to increase the discriminability of the shared Vector space.

\section{Data Used}
\label{sec:data}
Training and test data used in this work is the same as used in \cite{ali2016did}. Table~\ref{tab:data1} gives the number of hours of data available for each dialect for training and testing.

 \begin{table}[H]
 \centering 
 \begin{tabular}{@{}llllll@{}} \toprule Data & EGY & GLF & LAV & NOR & MSA \\ \midrule
 Train &  13 & 10 & 11 & 9 & 10 \\
Test & 2 & 2 & 2 & 2 & 2 \\

 \bottomrule \end{tabular}
 \caption{\textit{Number of hours of training and testing data for each dialect}}\label{tab:data1}
 \end{table}

Table~\ref{tab:data2} shows the number of speech utterances that are available for training and testing the DID system.
 \begin{table}[htb]
 \centering 
 \begin{tabular}{@{}llllll@{}} \toprule Data & EGY & GLF & LAV & NOR & MSA \\ \midrule
 Train &  1720 & 1907 & 1059 & 1934 & 1820 \\
Test & 315 & 348 & 238 & 355 & 265 \\
 \bottomrule \end{tabular}
 \caption{\textit{Number of training and test utterances for DID system development}}\label{tab:data2}
 \end{table}

\textbf{Training data} consist of recording from the Arabic Broadcast domain and contains utterances spoken in all the five dialects; EGY, GLF, LAV, MSA and NOR. 

The \textbf{test set} is from the same broadcast domain but is collected from Al-Jazeera and hence, unlike training data set, the recording are of high quality. The test set is labeled using CrowdFlower, a crowd source platform, by QCRI and is publicly available on their web portal\footnote{https://github.com/Qatar-Computing-Research-Institute/dialectID}. More details about the train and test data can be found in \cite{ali2016did,wray2015crowdsource}.
 
\begin{figure*}
  \centering
  \includegraphics[width=\linewidth]{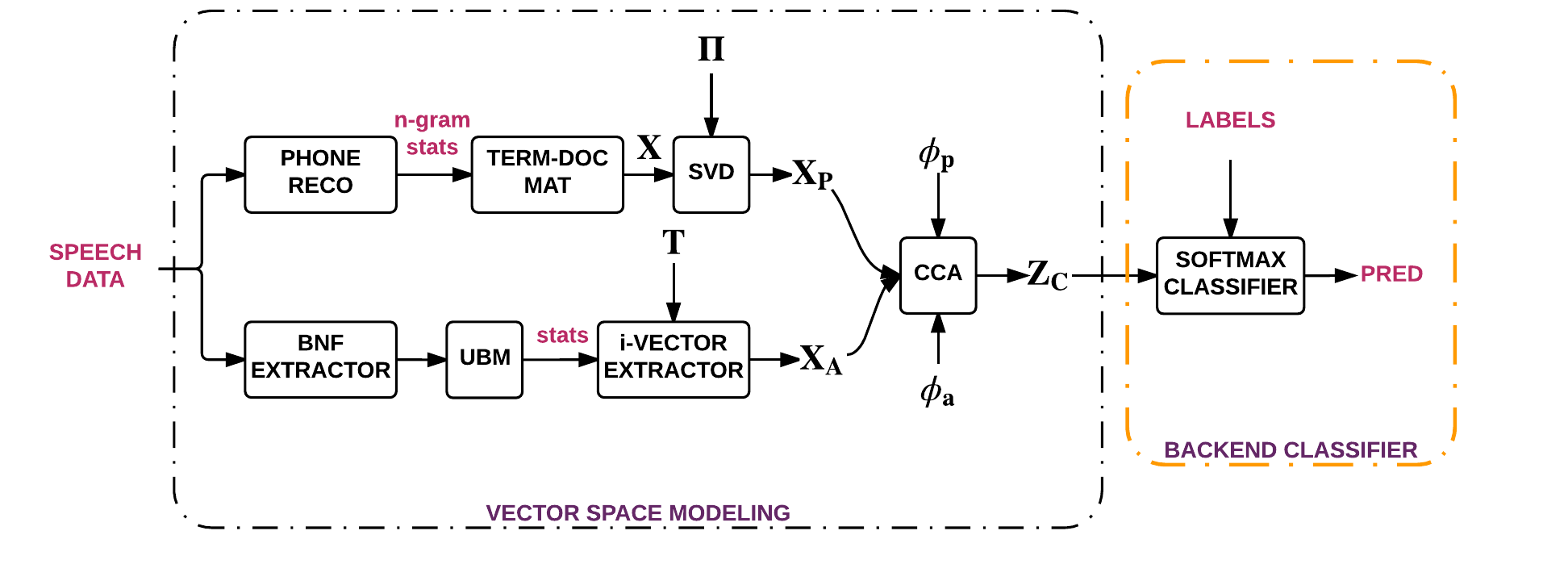}
  \caption{\textit{High level block diagram of our Dialect Identification System}}
  \label{sys}
\end{figure*}

\section{System Description}
\label{sec:sys}
Fig~\ref{sys} gives an overview of our DID system, which can be seen as a combination of two broad components; 1) Vector Space Modeling Component and 2) Back-end classifier. 

The most important pieces of the DID system are the four latent Vector Subspaces; 1) $\mathbf{\Pi}$: which is learned by performing SVD on the n-gram phootactic \textit{term-document} matrix (Section~\ref{sec:pvsm}), 2) $\mathbf{T}$: is the \textit{total variability subspace} that is learned in an unsupervised manner in the i-Vector framework (Section~\ref{sec:avsm}), 3) $\phi_{p}$: is the latent vector subspace corresponding to the Phonotactic VSM learned in the CCA Vector Space modeling framework and 4) $\phi_{a}$: is the latent vector subspace corresponding to Acoustic VSM learned in the CCA vector space modeling framework (Section~\ref{sec:ccavsm}). The shared VSM is then fed to a back-end discriminative classifier. In our case, we use \textit{multi-class logistic regression}, also known as \textit{softmax classification} for DID \cite{murphy2012machine,zadrozny2002transforming}. 

The parameter settings used for the \textit{softmax classifier} can be found in Table~\ref{tab:lr}. We use \textit{elastic net} regularization i.e. using both an L1 and L2 regularizer. \textit{Stochastic Gradient Descent} is used for learning the model parameters. \textit{Log-entropy} loss function is used as the model training objective.
\begin{table}[htb]
 \centering 
 \begin{tabular}{@{}lllll@{}} \toprule Model & L1-ratio & L2-ratio & loss & training \\ \midrule
 Softmax &  0.5 & 0.5 & log-entropy & SGD  \\
 \bottomrule \end{tabular}
 \caption{\textit{Parameters of the \textit{softmax classifier} model}}\label{tab:lr}
 \end{table}

\section{Experiments and Results}
\label{sec:exp}
\subsection{Dialect Identification Results}
Table~\ref{tab:res} gives performance of different VSMs on Arabic DID task. The standalone Phonotactic VSM, $\mathbf{X_{P}}$, performs rather poorly as compared to the standalone Acoustic VSM, $\mathbf{X_{A}}$. CCA VSM, $\mathbf{Z_{C}}$, which is formed by combining $\mathbf{X_{P}}$ and $\mathbf{X_{A}}$ using CCA performs better than both standalone VSMs, which is an expected outcome because $\mathbf{Z_{C}}$ carries dialect discriminative information from both $\mathbf{X_{P}}$ and $\mathbf{X_{A}}$. 

Further improvements in performance are due to performing LDA and WCCN on $\mathbf{Z_{C}}$. LDA is a supervised dimensionality reduction technique that form a lower dimensional Vector Subspace such that the class separability between the data points is maximized, and hence the performance gain that we see due to LDA is to be expected. An interesting observation to note is that the LDA based Acoustic VSM, $\mathbf{X_{A+LDA+WCCN}}$, performs at par with the LDA based CCA VSM, $\mathbf{Z_{C+LDA+WCCN}}$. This shows that some of the useful discriminative information from the Acoustic VSM is lost while performing CCA and hence we finally concatenate the two LDA based VSMs to get the final accuracy of 60\% on the DID task.

\begin{figure}[H]
  \centering
  \includegraphics[width=\linewidth]{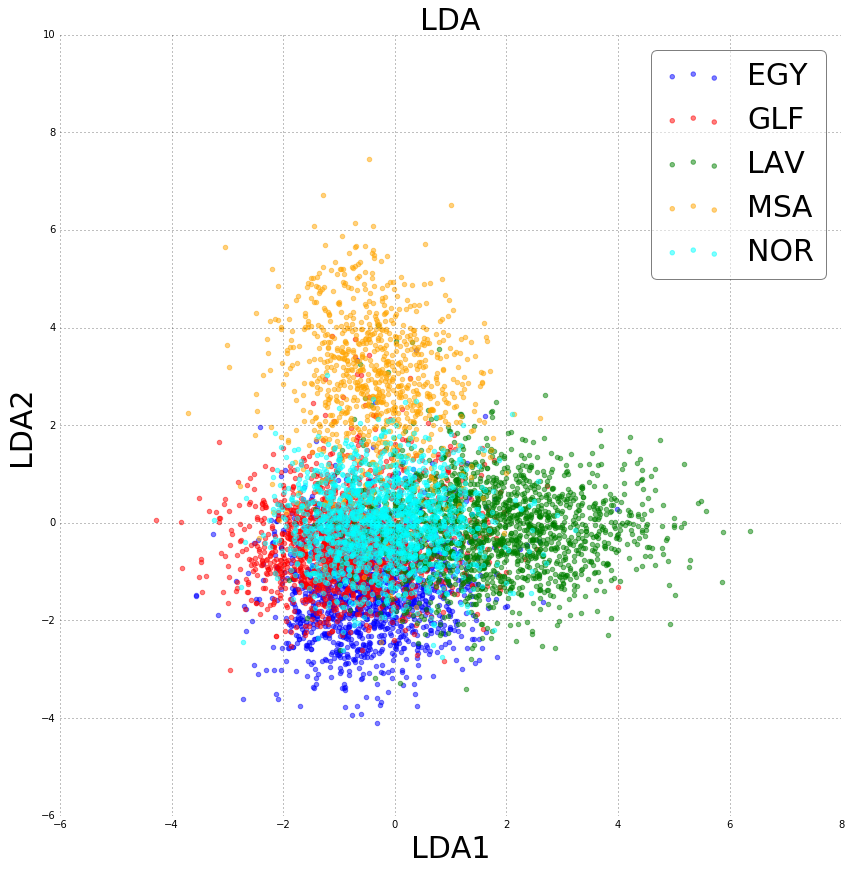}
  \label{proj}
  \caption{The first two components of the LDA based VSM used for DID}
\end{figure}

\begin{table}[htb]
 \centering 
 \begin{tabular}{@{}lllll@{}} \toprule VSM & d & ACC & PRC & RCL \\ \midrule
 $\mathbf{X_{P}}$ &  1200 & 0.45 & 0.45 & 0.46 \\
$\mathbf{X_{A}}$ & 400 & 0.55 & 0.61 & 0.55 \\
$\mathbf{Z_{C}}$ &  600 & 0.56 & 0.61 & 0.57  \\
$\mathbf{Z_{C+LDA+WCCN}} (\mathcal{A})$ &  4 & 0.58 & 0.62 & 0.58\\
$\mathbf{X_{A+LDA+WCCN}} (\mathcal{B})$ & 4 & 0.58 & 0.63 & 0.60 \\
$\mathcal{A} + \mathcal{B}$ & 8 & \textbf{0.60} & 0.63 & 0.60 \\
 \bottomrule \end{tabular}
 \caption{\textit{DID Results with Different VSMs. Accuracy, Precision and Recall}}\label{tab:res}
 \end{table}

\subsection{Confusion}
Table~\ref{tab:con} gives the confusion matrix for the Arabic DID task. We can infer the following: 1) EGY is confused most often with LAV, 2) GLF is confused most often with LAV and EGY, 3) LAV most often confused with EGY and GLF, 4) MSA is pretty well discriminated, which can also be confirmed by the VSM projection given by Fig~\ref{proj}, 5) NOR is most confused with EGY and LAV, which can also be seen in the VSM projection, where LAV is given by cyan region while EGY and LAV are given by blue and green dots.

 \begin{table}[H]
 \centering 
 \begin{tabular}{@{}llllll@{}} \toprule  & EGY & GLF & LAV & MSA & NOR \\ \midrule
 EGY &  \textbf{229} & 15 & 52 & 6 & 12 \\
GLF & 50 & \textbf{127} & 74 & 9 & 4 \\
LAV &  70 & 39 & \textbf{205} & 15 & 16  \\
MSA &  13 & 18 & 25 & \textbf{219} & 4\\
NOR & 81 & 26 & 78 & 11 & \textbf{158} \\
 \bottomrule \end{tabular}
 \caption{\textit{DID Results with Different VSMs. Accuracy, Precision and Recall}}\label{tab:con}
 \end{table}

\section{Conclusions}
\label{sec:conc}
In this work, we showed our innovative approach to construct a single VSM for DID that carries dialect discriminative information from both the Acoustic and Phonotactic VSMs. To that end, we use a well known multi-view dimensionality reduction known as Canonical Correlation Analysis (CCA). The single VSM constructed performed better than any of the Phonotcatic or Acoustic VSMs alone, but the LDA based CCA and Acoustic VSMs performed at par on the DID taks, while their combination gave us our best VSM for Arabic DID. We conclude that some dialect specific discriminative information is lost while performing CCA between Acoustic and Phonotactic VSMs and hence the final combination performs better. CCA is an unsupervised method and can easily incorporate unlabeled data from a different domain and act as a domain adaptation or semi-supervised learning method such as co-training as shown in \cite{foster2008multi}. We leave co-training and domain adaptation using CCA for our future work.

\bibliographystyle{IEEEbib}
\bibliography{strings}

\begin{thebibliography}{10}

\bibitem{biadsy2011automatic}
Fadi Biadsy,
\newblock {\em Automatic dialect and accent recognition and its application to
  speech recognition},
\newblock Ph.D. thesis, Columbia University, 2011.

\bibitem{ali2016did}
Ahmed Ali, Najim Dehak, Patrick Cardinal, Sameer Khurana, Sree~Harsha Yella,
  James Glass, Peter Bell, and Steve Renals,
\newblock ``Automatic dialect detection in arabic broadcast speech,''
\newblock in {\em Interspeech 2016}, 2016, pp. 2934--2938.

\bibitem{li2007vector}
Haizhou Li, Bin Ma, and Chin-Hui Lee,
\newblock ``A vector space modeling approach to spoken language
  identification,''
\newblock {\em IEEE Transactions on Audio, Speech, and Language Processing},
  vol. 15, no. 1, pp. 271--284, 2007.

\bibitem{deerwester1990indexing}
Scott Deerwester, Susan~T Dumais, George~W Furnas, Thomas~K Landauer, and
  Richard Harshman,
\newblock ``Indexing by latent semantic analysis,''
\newblock {\em Journal of the American society for information science}, vol.
  41, no. 6, pp. 391, 1990.

\bibitem{moore1981principal}
Bruce Moore,
\newblock ``Principal component analysis in linear systems: Controllability,
  observability, and model reduction,''
\newblock {\em IEEE transactions on automatic control}, vol. 26, no. 1, pp.
  17--32, 1981.

\bibitem{zissman2001automatic}
Marc~A Zissman and Kay~M Berkling,
\newblock ``Automatic language identification,''
\newblock {\em Speech Communication}, vol. 35, no. 1, pp. 115--124, 2001.

\bibitem{biadsy2009spoken}
Fadi Biadsy, Julia Hirschberg, and Nizar Habash,
\newblock ``Spoken arabic dialect identification using phonotactic modeling,''
\newblock in {\em Proceedings of the eacl 2009 workshop on computational
  approaches to semitic languages}. Association for Computational Linguistics,
  2009, pp. 53--61.

\bibitem{biadsy2011dialect}
Fadi Biadsy, Julia Hirschberg, and Daniel~PW Ellis,
\newblock ``Dialect and accent recognition using phonetic-segmentation
  supervectors.,''
\newblock in {\em INTERSPEECH}, 2011, pp. 745--748.

\bibitem{torresapproaches}
Pedro~A Torres-Carrasquillo, Elliot Singer, Mary~A Kohler, Richard~J Greene,
  Douglas~A Reynolds, and JR~Deller~Jr,
\newblock ``Approaches to language identification using gaussian mixture models
  and shifted delta cepstral features x, i,''
\newblock .

\bibitem{martinez2012ivector}
David Mart{\'\i}nez, Luk{\'a}{\v{s}} Burget, Luciana Ferrer, and Nicolas
  Scheffer,
\newblock ``ivector-based prosodic system for language identification,''
\newblock in {\em 2012 IEEE International Conference on Acoustics, Speech and
  Signal Processing (ICASSP)}. IEEE, 2012, pp. 4861--4864.

\bibitem{dehak2011language}
Najim Dehak, Pedro~A Torres-Carrasquillo, Douglas~A Reynolds, and Reda Dehak,
\newblock ``Language recognition via i-vectors and dimensionality reduction.,''
\newblock in {\em INTERSPEECH}. Citeseer, 2011, pp. 857--860.

\bibitem{ccahoteling}
H.~Hotelling,
\newblock ``Canonical correlation analysis (cca),''
\newblock in {\em Journal of Educational Psychology}, 1935.

\bibitem{cardinal2015speaker}
Patrick Cardinal, Najim Dehak, Yu~Zhang, and James Glass,
\newblock ``Speaker adaptation using the i-vector technique for bottleneck
  features,''
\newblock {\em Proceedings of Interspeech}, vol. 2015, 2015.

\bibitem{ali2014complete}
Ahmed Ali, Yifan Zhang, Patrick Cardinal, Najim Dahak, Stephan Vogel, and James
  Glass,
\newblock ``A complete kaldi recipe for building arabic speech recognition
  systems,''
\newblock in {\em Spoken Language Technology Workshop (SLT), 2014 IEEE}. IEEE,
  2014, pp. 525--529.

\bibitem{kenny2008study}
Patrick Kenny, Pierre Ouellet, Najim Dehak, Vishwa Gupta, and Pierre Dumouchel,
\newblock ``A study of interspeaker variability in speaker verification,''
\newblock {\em IEEE Transactions on Audio, Speech, and Language Processing},
  vol. 16, no. 5, pp. 980--988, 2008.

\bibitem{dhillon2015eigenwords}
Paramveer~S Dhillon, Dean~P Foster, and Lyle~H Ungar,
\newblock ``Eigenwords: Spectral word embeddings,''
\newblock {\em The Journal of Machine Learning Research}, vol. 16, no. 1, pp.
  3035--3078, 2015.

\bibitem{hardoon2003canonical}
David~R Hardoon, Sandor Szedmak, and John Shawe-Taylor,
\newblock ``Canonical correlation analysis; an overview with application to
  learning methods,''
\newblock 2003.

\bibitem{wray2015crowdsource}
Samantha Wray and Ahmed Ali,
\newblock ``Crowdsource a little to label a lot: Labeling a speech corpus of
  dialectal arabic,''
\newblock in {\em INTERSPEECH}, 2015.

\bibitem{murphy2012machine}
Kevin~P Murphy,
\newblock {\em Machine learning: a probabilistic perspective},
\newblock MIT press, 2012.

\bibitem{zadrozny2002transforming}
Bianca Zadrozny and Charles Elkan,
\newblock ``Transforming classifier scores into accurate multiclass probability
  estimates,''
\newblock in {\em Proceedings of the eighth ACM SIGKDD international conference
  on Knowledge discovery and data mining}. ACM, 2002, pp. 694--699.

\bibitem{foster2008multi}
Dean~P Foster, Sham~M Kakade, and Tong Zhang,
\newblock ``Multi-view dimensionality reduction via canonical correlation
  analysis,''
\newblock .

\end{thebibliography}
\end{document}